%% file: main.tex
\documentclass[sigconf]{acmart}
\pdfoutput=1

\usepackage{subfigure}
\usepackage{caption}
\usepackage{booktabs} 

\setcopyright{rightsretained}

\acmConference[CODS-COMAD 2019]{The ACM India Joint International Conference on
Data Science \& Management of Data}{January 2019}{Kolkata, West Bengal India}
\acmYear{2019}
\copyrightyear{2019}

\begin{document}
\title{Sequential Learning of Movement Prediction in Dynamic Environments using LSTM Autoencoder}

\author{Meenakshi Sarkar}
\affiliation{%
  \institution{PhD Student, Indian Institute of Science}
  \city{Bangalore}
  \state{Karnataka. India}
  \postcode{560012}
}
\email{meenakshisar@iisc.ac.in}

\author{Debasish Ghose}
\affiliation{%
  \institution{Professor, Indian Institute of Science}
  \city{Bangalore}
  \state{Karnataka, India}
  \postcode{560012}
}
\email{dghose@iisc.ac.in}

\begin{abstract}
Predicting movement of objects while the action of learning agent interacts with the dynamics of the scene still remains a key challenge in robotics. We propose a multi-layer Long Short Term Memory (LSTM) autoendocer network that predicts future frames for a robot navigating in a dynamic environment with moving obstacles. The autoencoder network is composed of a state and action conditioned decoder network that reconstructs the future frames of video, conditioned on the action taken by the agent. The input image frames  are first transformed into low dimensional feature vectors with a pre-trained encoder network and then reconstructed with the LSTM autoencoder network to generate the future frames. A virtual environment, based on the OpenAi-Gym framework for robotics, is used to gather training data and test the proposed network. The initial experiments show promising results indicating that these predicted frames can be used by an appropriate reinforcement learning framework in future to navigate around dynamic obstacles.
\end{abstract}
%

\ccsdesc[500]{Computing methodologies~Neural networks}

\keywords{Autoencoder, Unsupervised Learning, LSTM, Movement Prediction}

\maketitle

\input{other.tex}
\input{main_bib.bbl}

\end{document}

%% file: other.tex

\section{Introduction}

Prediction of movement of obstacles still remains a key challenge for robot path planning problems in dynamic environments.  For example, an Unmanned Aerial Vehicle (UAV) may fly in a cluttered environment, such as  a forest trail under high wind conditions, using a vision based path planning algorithm. Apart from the motion of the objects present in the environment, the motion of the UAV and the camera mounted on it also has an effect on the scene dynamics as perceived by the UAV.  Thus, the trajectory of the predicted future frames depends upon the action taken by the UAV at each time instant. In turn the decision of the path planning algorithm depends on the predicted future frames as these frames evolve according to the past actions taken by the agent.

%


Most existing techniques on learning to predict the dynamics of a scene depends upon hand labeled data. These techniques prove to be inadequate during unsupervised learning of environmental dynamics for training an agent autonomously. In such scenarios a possible approach could be to learn the representations of the scene and use that knowledge to make predictions about the future, conditioned on the action taken by the agent. Learning such models have proven to be a challenging task \cite{finn} because of the inherent complex and stochastic nature of the real world.

In the present work, we intend to explore the possibility of using LSTM networks for unsupervised learning  
in such scenarios. The encoding LSTM network maps the input
image frame sequence into a state vector of fixed length and the decoder network uses
this encoded information to reconstruct the input image frames and make a prediction about the
possible future states. These predicted frames in future could be used for successful path
planning in a fast changing environment.

\section{Motivation and related work}
Supervised learning of visual data representation has been quite successful in the past two decades. They not only learn good representation of the data set but also transfer well from one task to another and to different data sets. Researchers have tried to extend those same techniques to learn video data representations. This led to works that tried finding different ways to represent visual data using convolutional networks \cite{simonyan}. Unlike images, video data has issues of larger dimensionality. Learning representations of video data is challenging as it contains both spatial and temporal regularities. Thus, in order to learn long range relationships in the data, either hand engineered feature learning needs to be done or models need to be fed with a large number of labeled data. While costly labeled data with hand engineered kernel functions can successfully solve a particular problem, it fails to generalise. This justifies the need to develop methods to learn visual representation in an unsupervised manner. The work in \cite{ranzanto} proposes an autoencoder model which works well in learning visual representations of static images. A similar idea was used by  Srivastava et al. \cite{srivastava} to design an \textit{Encoder-Decoder} model for video data representation using LSTM networks.

LSTM networks have lately gained popularity among ML researchers because of their successful performance in sequence learning tasks such as speech and hand writing recognition \cite{graves}. These performances were achieved in a supervised learning paradigm. In their sequence to sequence language translation model Sutskever et al. \cite{sutskever}  used a recurrent neural network as an encoder that represented the word embedding of one language and another recurrent neural network that decoded those embedding vectors and translated them to another language. The LSTM \textit{Encoder-Decoder} model by \cite{srivastava} is inspired by the work of \cite{sutskever},and used one layer of LSTM units  as an encoder that encoded and learned the representation of video data and the second LSTM layer decoded those feature vectors and reconstructed the input image sequence. A similar architecture could be used to predict future frames also. 

In our proposed approach we use an \textit{Encoder-Decoder} model similar to \cite{srivastava} but instead of feeding the LSTM network with the raw image sequences we first transform the sequence of images into a sequence of feature vectors by passing them  through a pre-trained encoder layer. At the output layer, we pass the reconstructed feature vectors generated by the decoder LSTM layer through a pre-trained decoder layer in order to reconstruct the input sequence. 

We further propose an action and state conditioned LSTM Autonecoder network that would capture the physical interactions of the agent with the environment and predict the future, conditioned on the action taken by the agent. Recently Finn et al. \cite{finn} have come up with a similar conceptual model. Their model is based on convolutional LSTM networks and is computationally taxing as they propose to capture the dynamics of the interactions of the agent with the objects present in the scene via conditioning the transformation of each pixel in the image based on the information available on the action taken by the agent and the most recent past. In our work we attempt to extend the idea of state and action conditioned LSTM network to solve the path planning problem of avoiding dynamic obstacles. To our knowledge this is the first attempt made to address a path planning problem with moving obstacles. In \cite{inoue} an LSTM autoencoder network was used for point to point navigation in a constraint environment but with static obstacles. Unlike \cite{finn} our smaller parametric model of two layer LSTM network does not suffer from computational overloads. We further reduce the required number of estimated parameters by embedding the features of raw image by passing it through a pre-trained dense layer. The LSTM layer then learns the temporal features from the data in the mapped feature space.

The contribution of this paper is summarized as follows: $(i)$ extension of the concept of state and action conditioned LSTM autoencoder networks to solve robot path planning problems in dynamic environment; $(ii)$ reduction of the size of the parametric model of the LSTM network with the introduction of the pre-trained layers for mapping the high dimensional image input to a low dimensional feature space. 

\section{Model Description}
The base unit of the \textit{Encoder-Decoder} network is the LSTM cell block. We use the model given by \cite{graves} to implement our model.
\subsection{Long Short Term Memory (LSTM) Cell}
The LSTM cell (Figure \ref{fig1}) is the basic building block of our model. Each LSTM cell has a state at time $t$, denoted as $\mathbf{c}_t$. The cell acts as a memory unit and the access for reading or modifying the contents of this unit is controlled through three sigmoidal gates: input gate $\mathbf{i}_t$, forget gate $\mathbf{f}_t$ and output gate $\mathbf{o}_t$. At
each time step the unit receives inputs from two external sources (the current frame $\mathbf{x}_t$ and the previous hidden states of all LSTM units in the same layer $\mathbf{h}_{t-1}$) at each of the three gates and the input. Additionally, each gate has a peephole connection to  internal cell state $\mathbf{c}_{t-1}$ of its own cell block. The
inputs coming from different sources are summed along with a bias. The gates use logistic activation functions. First the total input at the input terminal passes through a non-linear function like tanh and then it is multiplied with the activation of the input gate. The resulting outcome is then added to the cell state of the previous time step $\mathbf{c}_{t-1}$ after the cell state is multiplied with the activation of the forget gate $\mathbf{f}_t$. The outcome is assigned as the new cell state $\mathbf{c}_t$ of the unit. The final output from the LSTM unit $\mathbf{h}_t$ is computed after passing the cell state through a non-linearity like tanh and multiplying the outcome with the activation of the output gate $\mathbf{o}_t$. The updates are summarized as follows \cite{graves} :
\begin{figure}
    \centering
    \includegraphics[width=.3\textwidth]{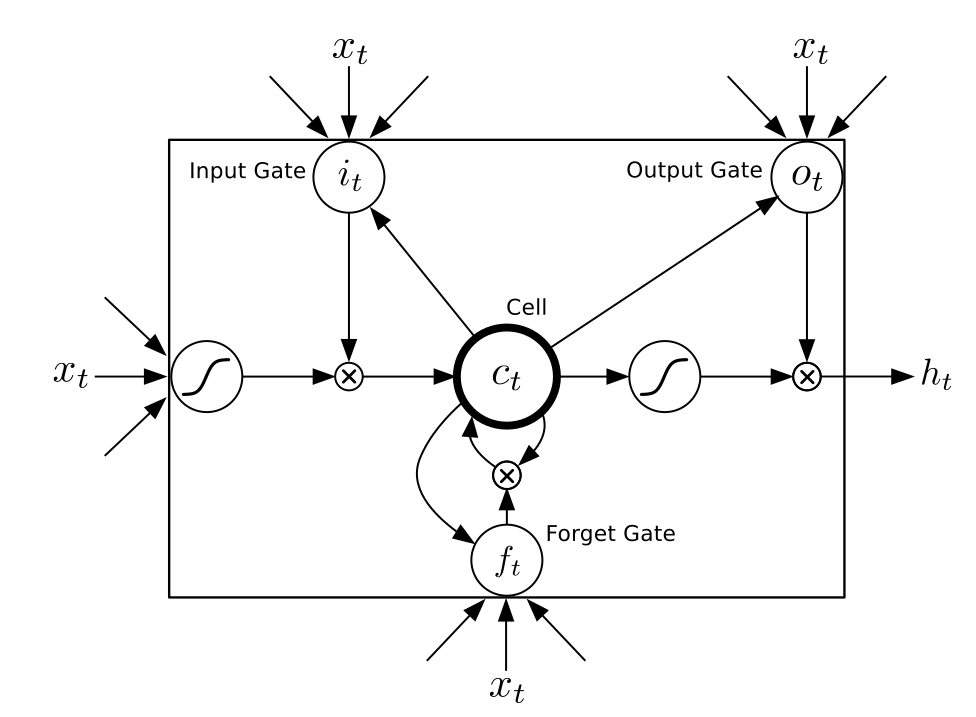}
    \caption{Architecture of LSTM Unit}
    \label{fig1}
\end{figure}
\begin{equation*}
    \centering
  \mathbf{i}_t=\sigma (W_{xi}\mathbf{x}_t+W_{hi}\mathbf{h}_{t-1}+W_{ci}\mathbf{c}_{t-1}+\mathbf{b}_i)
  \end{equation*}
  \begin{equation*}
  \centering
  \mathbf{f}_t=\sigma (W_{xf}\mathbf{x}_t+W_{hf}\mathbf{h}_{t-1}+W_{cf}\mathbf{c}_{t-1}+\mathbf{b}_f)
  \end{equation*}
  \begin{equation*}
  \centering
  \mathbf{c}_t=\mathbf{f}_t\mathbf{c}_{t-1}+\mathbf{i}_{t}\mathrm{tanh}(W_{xc}\mathbf{x}_t+W_{hc}\mathbf{h}_{t-1}+\mathbf{b}_c)
  \end{equation*}
  \begin{equation*}
  \centering
  \mathbf{o}_t=\sigma (W_{xo}\mathbf{x}_t+W_{ho}\mathbf{h}_{t-1}+W_{co}\mathbf{c}_{t}+\mathbf{b}_o)
  \end{equation*}
  \begin{equation*}
  \centering
  \mathbf{h}_t=\mathbf{o}_t\mathrm{tanh}(\mathbf{c}_t)
\end{equation*}

All the weight matrices except the diagonal $W_{c\cdot}$ are of dense form. Unlike recurrent neural networks LSTM does not suffer from the vanishing gradient problems. Thus, it is much easier to learn long term regularities among a sequence of data using LSTM network.

\subsection{LSTM Encoder-Decoder Model}
       
\begin{figure}[h!]
        \centering
        \includegraphics[width=.45\textwidth]{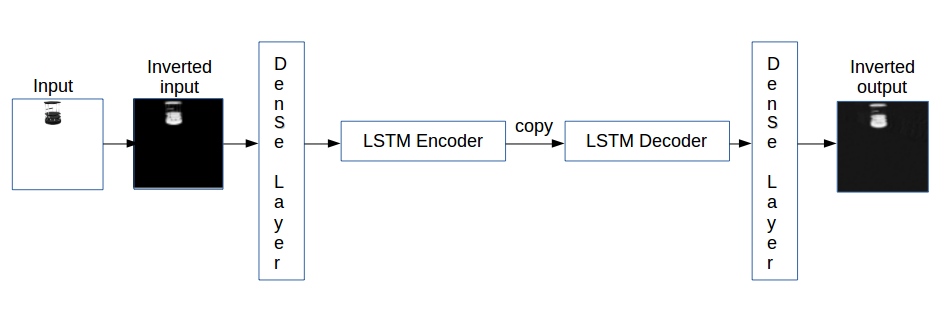}
       
        \caption{Schematic representation of the complete LSTM Autoencoder Network}
        \label{fig3}
\end{figure}

The main Encoder-Decoder model consists of two LSTM layers for the unsupervised learning. Figure \ref{fig3} shows a schematic diagram of the LSTM autoencoder network used for our applications. Instead of feeding the raw images to the RNN units we first pass the images through a pre-trained fully connected layer of 128 neurons. These neurons are used to learn the essential spatial feature representations from the data. The dense layer transforms the sequence of raw images to a sequence of feature vectors. These vectors are then fed to the encoding LSTM layer. The LSTM layer further captures the temporal and spatial regularities present in the sequence of images. There were two decoder modules for the LSTM network. One was used to reconstruct the input image sequence and another to make prediction of the future frames. In both cases the decoder module consists of only one LSTM layer. The decoder unit reconstructs the temporal and spatial regularities learned from the encoder module and the decoded sequence of feature vectors passes through a dense reconstruction layer to make predictions about future frames or get back the original input image sequence. The inversion in the input and output was found to be necessary and will be explained later.

For the predictive network we input the frames of a sequence to the encoder one after another and let it learn the regularities in the data and once it does so, the final representation is used to predict the future frames. For reconstruction module we tried two approaches: one which is similar to the one mentioned earlier and in the second approach the network learns and makes prediction after each frame of the sequence. We have used a mean square loss function for both the future frame prediction and input reconstruction models.

\subsection{Proposed Action and State Conditioned LSTM Autoencoder Model}
\begin{figure}
    \centering
    \includegraphics[width=.45\textwidth]{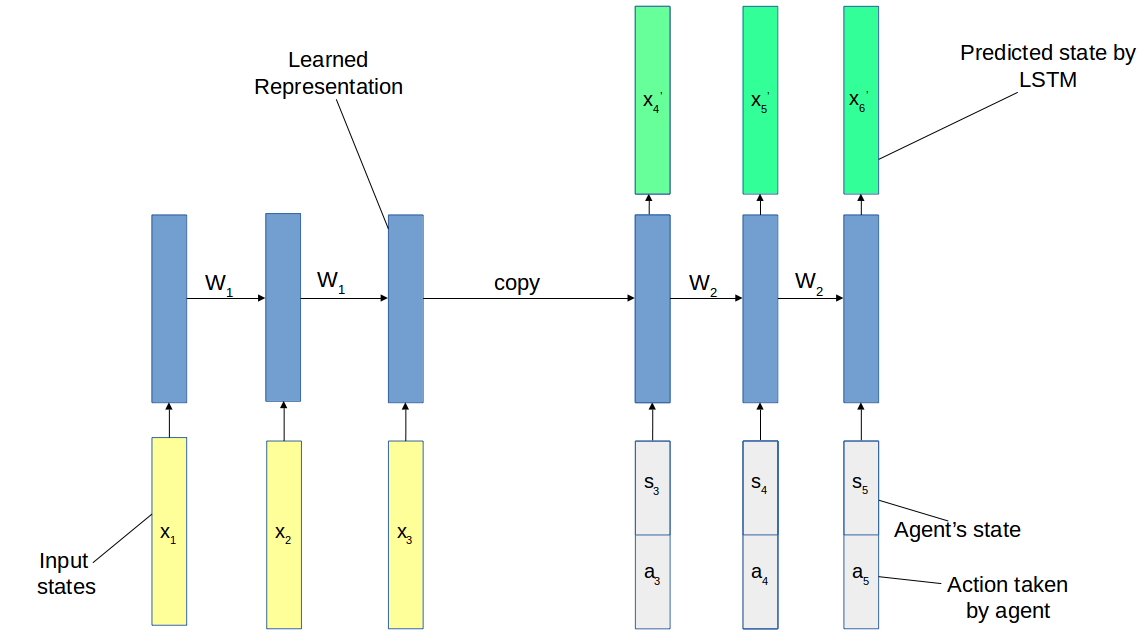}
    \caption{Schematic representation of a state and action conditioned LSTM autoencoder network}
    \label{fig4}
\end{figure}
Figure \ref{fig4} shows a conceptual model of our proposed state and action conditioned LSTM autoencoder network.  In their work  Srivastava et al. \cite{srivastava} mentioned the possibility of a  state conditioned LSTM predictor model and stated that state conditioned decoder might be able to model multi-modal target sequences. The non conditional models tend to average out the multiple modes in a low level input space. Thus, in order to capture the multiple modes of the target sequence, we use a  state conditioned model. This would help us model the physics of the interaction of the agent with the environment. We further propose that in order to select from all possible future trajectories, the decoder network must be conditioned on the action taken by the agent at time $t$. Thus, we further add the action tuple to the state vectors.


The paper \cite{finn} also talks about a model built on a similar philosophy where they predicted future frames conditioned on the present state and action taken by the agent. But the model in \cite{finn} suffers from computational overloads as it predicts the motion of each pixel in the scene while learning some notion of detecting objects as it tends to group together the pixels from a same object.  The model in \cite{finn} was inspired by the stacked convolutional LSTM networks given by \cite{xing}. In this work, we are extending the concept of state and action conditioned LSTM networks to the robot motion planning problem in dynamic environment. But instead of using convolutional LSTM networks like \cite{finn}, we are using LSTM autoencoder network to learn the temporal regularities of the fearure vectors extracted from the image frames. These low dimensiona feature vectors that capture the spatial regularities of the image are extracted  with a pre-trained encoder layer. Instead of focusing on learning the motion of each pixel in the scene our model learns the temporal regularities of the principal spatial features of the scene, using the LSTM autoencoder architecture where the future is conditioned on the action taken by the agent Our model is focused on learning. Thus, making it more computationally efficient that \cite{finn}'s model. We use a mean square loss objective function to optimize the network. Although we experimented with a few other  cost functions but for the application considered in this paper, mean square loss function seemed to yield the best results.

\section{Experimental Setup, Results and Analysis}
\begin{figure}[t]
        \centering
        \includegraphics[width=.45\textwidth]{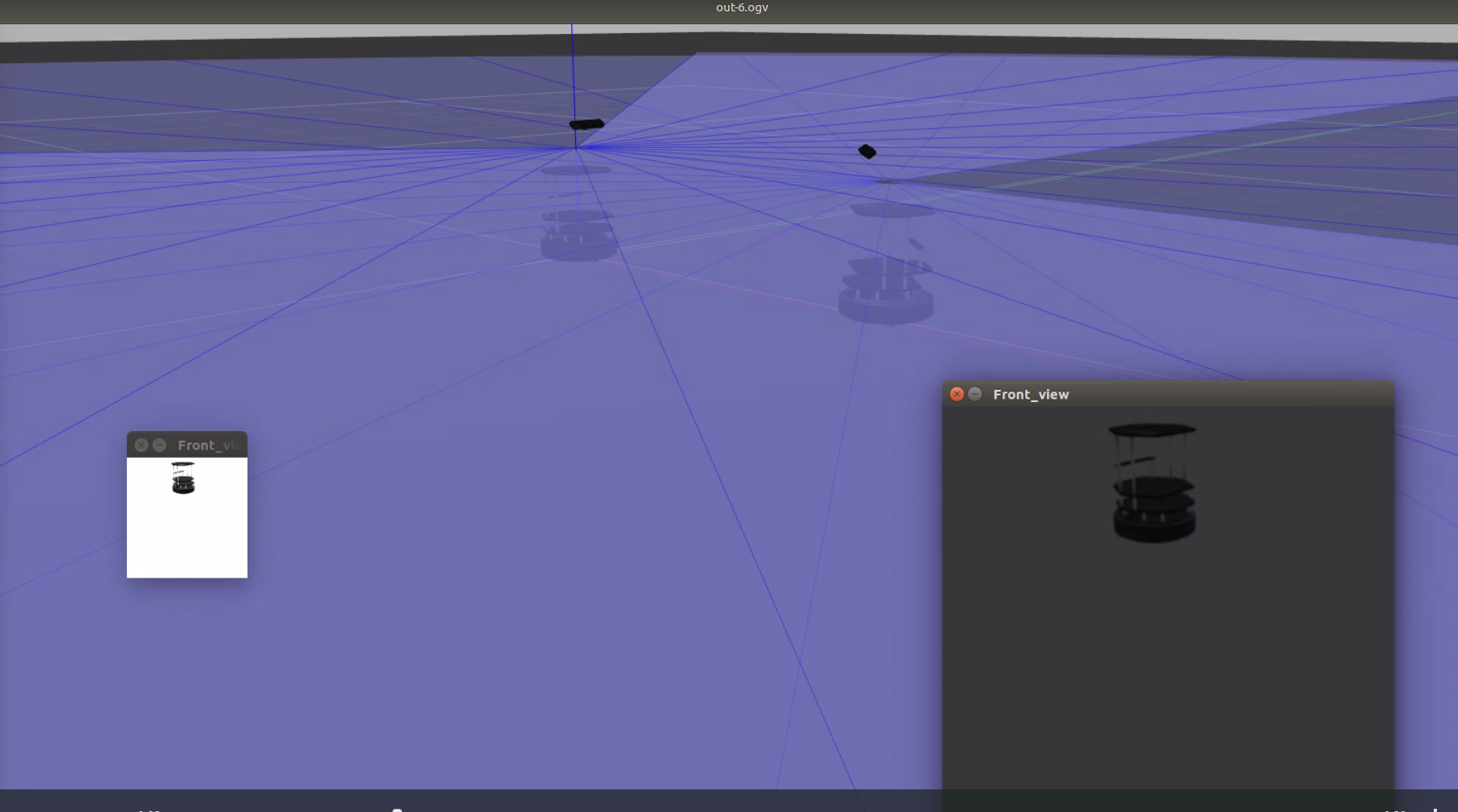}
       
        \caption{A ROS-Gazebo based virtual experimental environment has been set up in the OpenAI-gym framework.}
        \label{fig5}
    \end{figure}

Figure \ref{fig5} shows a snapshot of the ROS-Gazebo based virtual experimental environment that has been set up in the OpenAi-Gym framework for robotics \cite{openai}, to obtain training data for the LSTM network. The initial virtual setup has two turtlebots, Tb1 and Tb2. During the data collection phase, Tb1 is stationary while tracking the movement of Tb2 moving in front of it. Tb2 is made to move from point A to point B using a Proportional Integral Derivative (PID) controller. To introduce variations in the training data, during each trial of the 5000 training trials, Tb2 is controlled with different PID parameters and is made to go to 4 different end points from 3 different starting points.  
\begin{figure}[h!]
        \centering
        \includegraphics[width=.45\textwidth]{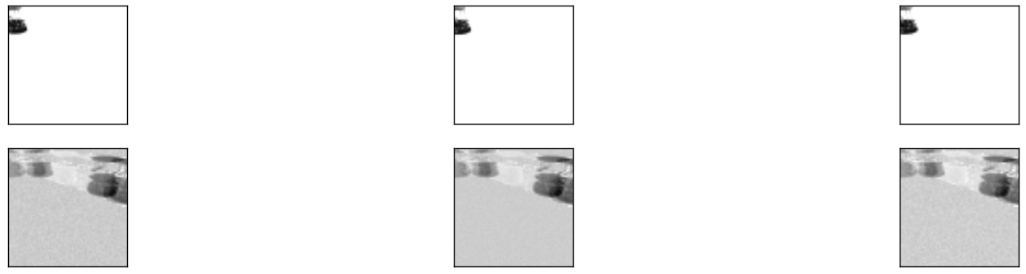}
       
        \caption{Reconstruction of input without inversion of input image $t-4$, $t-3$, $t-2$, $t-1$ and $t^{th}$ instance.}
        \label{fig6}
    \end{figure}
    
    \begin{figure}[h!]
        \centering
        \includegraphics[width=.45\textwidth]{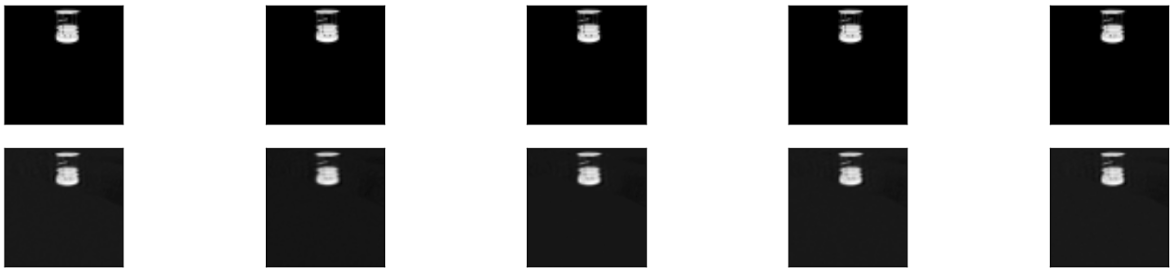}
     
        \caption{Reconstruction of input after inversion of input image for $t-4$, $t-3$, $t-2$, $t-1$ and $t^{th}$ instance. The first row of frames represents the ground truth and the last row represents the reconstructed frames by LSTM.}
        \label{fig7}
    \end{figure}
Initially, we tried to train the network with sequences of images that were transformed into a gray scale image after being smoothed through a Gaussian filter. We found that although the network learns some temporal regularities in the data, it fails to reconstruct the spatial features (see figure \ref{fig6}). This problem arose due to the highly sparse nature of the data and as the gradient values of the features (region) of interests were very small. Thus, the stochastic descent algorithm was not being able to update the parameters correctly. To avert this situation we inverted the images (see figure \ref{fig3}) and the results are shown in figure \ref{fig7}. The inversion operation is simply reassigning $\mathbb{I}_{(i,j)}= 255-\mathbb{I}_{(i,j)}$, where $\mathbb{I}_{(i,j)}$ represents the intensity value of a pixel and $i$ varies from $1$ to $m$ and  $j$ varies from $1$ to $n$.

\begin{figure}
        \centering
        \subfigure[]{
        \includegraphics[width=.45\textwidth]{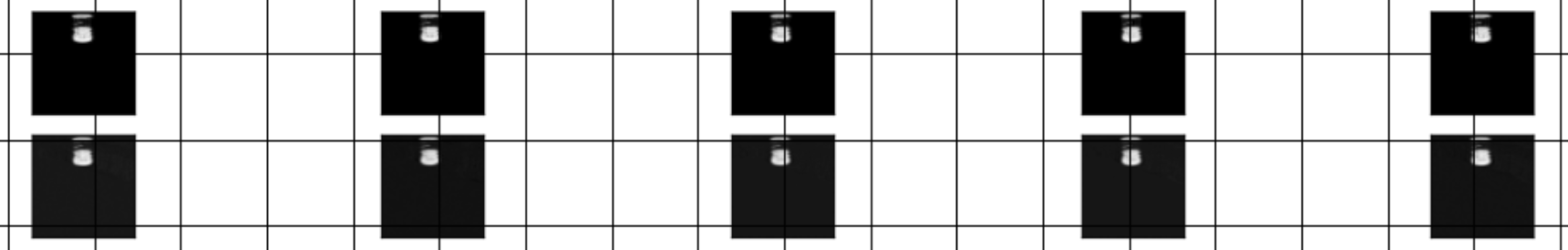}}\label{fig8a}
        
        \subfigure[]{
        \includegraphics[width=.45\textwidth]{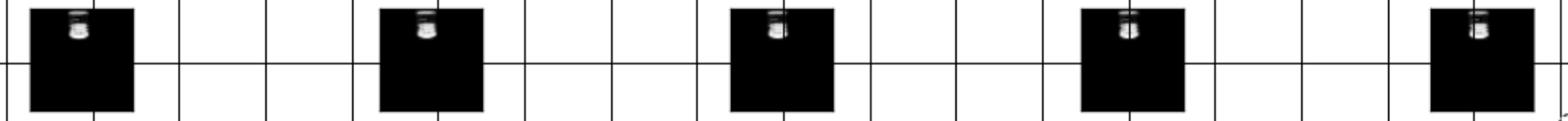}}\label{fig8b}
        \caption{The top row of Fig. 7a shows the predicted future frames at $(t+1)$, $(t+2)$, $(t+3)$, $(t+4)$ and $(t+5)^{th}$ time instances by LSTM with respect to the ground truth in the bottom row. Fig. 7b shows the input frames at  $(t-4)$, $(t-3)$, $(t-2)$, $(t-1)$ and $t^{th}$ time instance to the network. A careful examination of the vertical grids shows how the robot has moved with respect to time.}
        \label{fig8}
    \end{figure}

After successfully reconstructing the input sequences we have tried to make prediction about future frames and the results are shown in figure \ref{fig8}. Because of the smaller parametric space of our proposed network we observe that it can successfully make  prediction up to 5 frames in future after being trained on only 5000 frames which consists of 1000 sequences, each sequence having 5 frames. Although during the initial phase of the training we have kept the destination point of Tb2 same while the trajectories meant to arrive at the goal are different.    

\section{Conclusion and Future Work}

An OpenAI-gym and ROS framework \cite{openai} based virtual experimental setup has been created to test robot path planning algorithms in dynamic environment. For the first time  LSTM autoencoder network has been used to predict future frames for a robot navigating in a dynamic environment. These predicted future frames can be used to aid the path planning algorithm designed in a appropriate reinforcement learning framework to avoid moving obstacles in an unknown terrain. Introduction of the dense layer before the LSTM encoder unit has proven to make learning more efficient by mapping the high dimensional input space into a low dimensional feature space. 

As an ongoing work, We are yet to test the conceptual model of state and action conditioned LSTM autoencoder network for prediction of scenes with moving agent. Once the proposed model is tested, more insights can be obtained about whether or not LSTM networks would be able to capture the dynamics of the interaction between the scene and the action taken by the agent. \\